\documentclass{article}
\usepackage{spconf,amsmath,graphicx}

\usepackage[paper=letterpaper,
            left=0.63in,
            right=0.63in,
            top=1in,
            bottom=1.1in]{geometry}

\usepackage{enumitem}
\setlist{nosep, leftmargin=14pt}

\usepackage{mwe} 

\usepackage{spconf,amsmath,graphicx,hyperref}
\usepackage{caption}   
\usepackage{multirow}
\usepackage{booktabs}       
\usepackage{bbding}
\usepackage{xcolor}         
\usepackage{tabularx}  
\usepackage{makecell}  
\usepackage{wrapfig}
\usepackage{float} 
\usepackage{amsfonts}
\usepackage{cite}
\usepackage{etoolbox}
\apptocmd{\thebibliography}{\setlength{\itemsep}{0pt}\setlength{\parskip}{0pt}}{}{}

\usepackage{titlesec}
\titlespacing*{\section}{0pt}{*1}{*0.5}
\titlespacing*{\subsection}{0pt}{*0.8}{*0.3}
\title{SemiETPicker: Fast and Label-Efficient Particle Picking for CryoET Tomography Using Semi-Supervised Learning}%
\name{Linhan Wang$^{\star}$  
      Jianwen Dou$^{\star}$ 
      Wang Li$^{\dagger}$ 
      Shengkun Wang$^{\star}$ 
      Zhiwu Xie$^{\ddagger}$ 
      Chang-Tien Lu$^{\star}$ 
      Yinlin Chen$^{\star}$
      }

\address{$^{\star}$ Virginia Tech 
         $^{\dagger}$ University of Memphis 
         $^{\ddagger}$ Wenzhou-Kean University}
\begin{document}
%
\maketitle
\begin{abstract}
Cryogenic Electron Tomography (CryoET) combined with sub-volume averaging (SVA) is the only imaging modality capable of resolving protein structures inside cells at molecular resolution. Particle picking, the task of localizing and classifying target proteins in 3D CryoET volumes, remains the main bottleneck. Due to the reliance on time-consuming manual labels, the vast reserve of unlabeled tomograms remains underutilized. In this work, we present a fast, label-efficient semi-supervised framework that exploits this untapped data. Our framework consists of two components: (i) an end-to-end heatmap-supervised detection model inspired by keypoint detection, and (ii) a teacher–student co-training mechanism that enhances performance under sparse labeling conditions. Furthermore, we introduce multi-view pseudo-labeling and a CryoET-specific DropBlock augmentation strategy to further boost performance. Extensive evaluations on the large-scale CZII dataset show that our approach improves F1 by 10\% over supervised baselines, underscoring the promise of semi-supervised learning for leveraging unlabeled CryoET data.
\end{abstract}

\section{Introduction}
\label{sec:intro}

Cryogenic Electron Tomography (CryoET) enables the visualization of macromolecular complexes in near-native conformations at sub-nanometer resolution~\cite{doerr2017cryo}. In a typical CryoET pipeline, frozen hydrated samples are imaged by tilting them incrementally under an electron beam to acquire a tilt-series of 2D projections. These projections are computationally reconstructed into a 3D density map known as a tomogram. Downstream structural analysis critically relies on particle picking, i.e., the localization and classification of sub-cellular components within the tomogram. Existing approaches~\cite{wagner2019sphire, hao2022vp, gubins2020shrec, zhou2023machine} typically rely on supervised learning or template matching~\cite{frangakis2002identification}, both of which require extensive manual annotations. To alleviate annotation costs, recent efforts have explored few-shot learning~\cite{zhou2021one} and contrastive learning~\cite{huang2022accurate}. However, the wealth of unlabeled CryoET tomograms remains under-utilized.

Semi-Supervised Learning (SSL) leverages a small labeled dataset and a large unlabeled dataset to improve model performance. Landmark methods in image classification, such as Mean Teacher~\cite{tarvainen2017mean} and FixMatch~\cite{sohn2020fixmatch}, demonstrate that enforcing prediction consistency between weak and strong augmentations using high-confidence pseudo labels can yield substantial gains. In particular, FixMatch simplifies the SSL pipeline by employing the teacher model to generate pseudo labels from weakly augmented images, which are then used to supervise the student model trained on strongly augmented versions. Only pseudo labels with high confidence are retained, making this a simple and effective strategy in modern SSL.

Extending SSL to object detection introduces additional challenges due to class imbalance and the dual tasks of classification and localization~\cite{sohn2020simple, xu2021end, liu2021unbiased}. While particle picking bears similarities to natural image object detection, it presents unique difficulties: (1) the low signal-to-noise ratio in tomograms renders conventional deep architectures suboptimal; (2) confidence-based pseudo-label filtering alone fails to ensure quality supervision; and (3) strong augmentations effective for CryoET images, characterized by small, densely packed particles, are underexplored.

To address these challenges, we propose \textbf{SemiETPicker}, a novel semi-supervised particle picking algorithm tailored for CryoET. First, we design an efficient asymmetric U-Net architecture supervised by Gaussian heatmaps and optimized using a reweighted MSE loss. To replace complex non-maximum suppression (NMS), we introduce a lightweight max-pooling-based postprocessing module. This yields a fast and accurate end-to-end pipeline. We then adopt a teacher-student co-training framework where the teacher processes weakly augmented inputs to produce pseudo labels, and the student learns from strongly augmented images and those labels. To improve pseudo label reliability, we introduce a multi-view pseudo labeling strategy: the teacher generates predictions across multiple weak augmentations, and the average is used as the final pseudo label. Finally, we propose a CryoET-specific DropBlock augmentation, which perturbs the image while preserving particle structures, thereby providing effective supervision for densely packed regions.

Our contributions are summarized as follows: \textbf{1) We propose an efficient and accurate asymmetric U-Net architecture for particle picking}, supervised by Gaussian heatmaps with reweighted MSE loss, and postprocessed via an efficient max-pooling operation, enabling streamlined end-to-end inference. \textbf{2) We develop a teacher-student co-training semi-supervised learning strategy} built upon FixMatch and Mean Teacher principles, effectively leveraging unlabeled CryoET data. \textbf{3) We introduce two novel techniques to boost SSL performance}: (i) a multi-view pseudo labeling scheme to reduce uncertainty, and (ii) a CryoET-specific DropBlock augmentation designed for densely packed macromolecular structures.
\section{Methodology}
\label{sec:method}

Our SemiETPicker is a semi-supervised object detector that leverages unlabeled data to boost performance. We first describe the problem \ref{3_1}, then present the end-to-end detector \ref{3_2}. Section \ref{3_3} details the teacher–student co-training pipeline with multi-view pseudo-labeling, and Section \ref{3_4} introduces the CryoET-specific DropBlock augmentation.

\subsection{Problem Description}
\label{3_1}


Currently, a large number of CryoET tomograms have been generated. However, each tomogram contains hundreds of protein particles. Labeling such densely distributed and small particles is very time-consuming, so only a small fraction are annotated ~\cite{peck2024annotating}. We denote the labeled dataset as \(\mathcal{D}^L = \{(\mathbf{x}_i, y_i)\}_{i=1}^l\), where \(l \ll u\), and evaluation is performed on separate tomograms. The unlabeled dataset is denoted as \(\mathcal{D}^U = \{(\mathbf{x}^u_i)\}_{i=1}^u\), where \(\mathbf{x}_i \in \mathbb{R}^{D \times H \times W}\) are tomograms and \(y\) are labels. In the semi-supervised setting, we train our model jointly on \(\mathcal{D}^L\) and \(\mathcal{D}^U\).

\begin{figure}[t]
  \footnotesize
  \centering
  \includegraphics[width=0.476\textwidth]{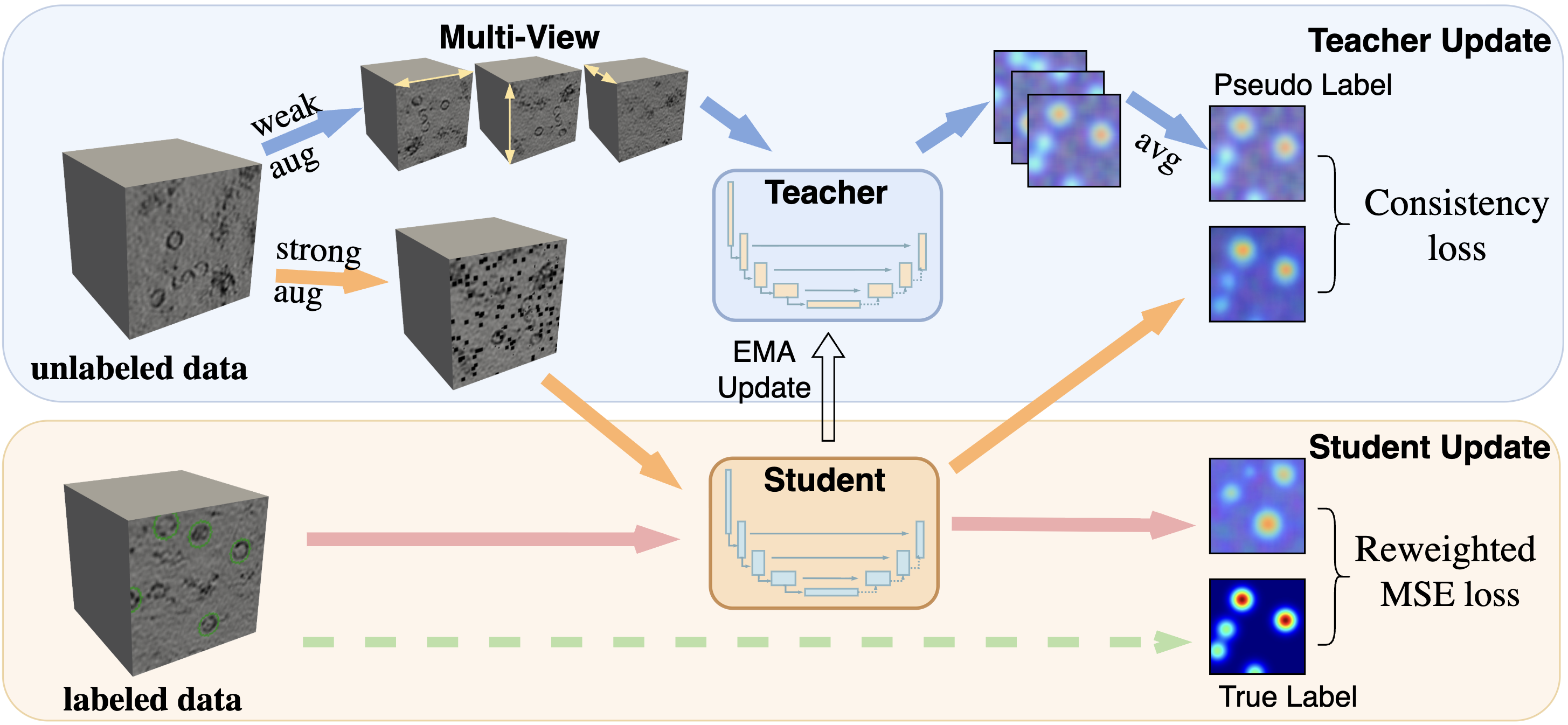}
  \caption{Overview of the SemiETPicker pipeline. The student model is jointly trained on labeled and unlabeled data using a combination of reweighted MSE loss and consistency loss. To reduce pseudo label uncertainty, we generate multi-view samples of unlabeled images via multiple flipping for the teacher model. The teacher is updated as the exponential moving average (EMA) of the student, enabling mutual improvement through a co-training loop.}
  \label{fig:obstruction_model}
  \vspace{-15pt}
\end{figure}

\subsection{An End-to-End Object Detector for Particle-Picking}
\label{3_2}

Here we reformulate the particle picking problem as a keypoint prediction problem inspired by~\cite{law2018cornernet, zhou2019objects}. Let \(I \in \mathcal{R}^{D \times H \times W}\) be an input tomogram of depth \(D\), height \(H\), and width \(W\). We aim to predict a heatmap \(\hat{Y} \in [0, 1]^{C \times \frac{D}{R} \times \frac{H}{R} \times \frac{W}{R}}\), where \(R\) is the output stride and \(C\) is the number of particle types. We experimentally find \(R = 2\) achieves a good trade-off between performance and speed. We design a asymmetric UNet accordingly to match the output size ~\cite{ronneberger2015u}.

Each label \(y_i\) contains a set of particle center coordinates and classes \(\{(p_j, c_j)\}_{j=1}^n\). For each center \(p \in \mathcal{R}^3\) of class \(c\), we compute its low-resolution equivalent \(\Tilde{p} = \left\lfloor \frac{p}{R} \right\rfloor\). We then splat all centers onto a heatmap \(Y \in [0, 1]^{C \times \frac{D}{R} \times \frac{H}{R} \times \frac{W}{R}}\) using a Gaussian kernel:

$Y_{czyx} = \exp\left( - \frac{(x - \Tilde{p}_x)^2 + (y - \Tilde{p}_y)^2 + (z - \Tilde{p}_z)^2}{2\sigma_c^2} \right)$,

where \(\sigma_c\) is the standard deviation, set to half the typical radius of particles of class \(c\). MSE loss is a common choice for heatmap regression~\cite{luo2021rethinking, uchida20252}. Due to the predominance of background voxels in CryoET volumes, we propose a reweighted MSE loss to address the severe class imbalance:

$\mathcal{L}_{hm}(\hat{Y}, Y) = \frac{\sum Y(\hat{Y} - Y)^2}{\sum Y + \epsilon} + \lambda \frac{\sum(1 - Y)(\hat{Y} - Y)^2}{\sum(1 - Y) + \epsilon}$

where \(\lambda = 4\) is a weighting factor. For clarity, we omit the subscripts \(czyx\) in the equation. Empirically, we find that using a larger \(\lambda\) improves performance, likely due to its role in emphasizing hard negative examples~\cite{shrivastava2016training}.

Since the model outputs heatmaps, we apply non-maximum suppression (NMS) via max pooling, using a kernel size equal to half the typical particle radius. After max pooling, we filter predictions by a confidence threshold; positions above the threshold are considered predicted particle centers. This pipeline is highly efficient, capable of processing a tomogram in under one second.

\subsection{Teacher-Student Co-training and Multi-View Pseudo Labeling}
\label{3_3}

Our semi-supervised learning pipeline consists of two stages. The first stage is the burn-in stage. We denote our model as \(f(x, \theta)\), where \(f\) represents an asymmetric U-Net architecture, \(x\) is the input tomogram, and \(\theta\) denotes the model weights. In this stage, we train the model solely on the labeled dataset \(\mathcal{D}^L\) to obtain initial weights \(\theta\).

The second stage is a teacher-student co-training stage. We duplicate the weights \(\theta\) to initialize both the teacher model \(\theta_t\) and the student model \(\theta_s\). The student model is trained using both labeled data \(\mathcal{D}^L\) and unlabeled data \(\mathcal{D}^U\). Meanwhile, the teacher model is updated via Exponential Moving Average (EMA) of the student weights:
$
\theta_t \leftarrow (1 - \alpha)\theta_t + \alpha \theta_s,
$
where \(\alpha\) is a smoothing coefficient. As demonstrated in prior work~\cite{tarvainen2017mean}, the EMA-updated teacher consistently outperforms the student model in prediction quality and is thus used to generate supervision signals.

To incorporate unlabeled tomograms, a consistency-based loss function is designed: the student should produce similar predictions as the teacher, even under strong input augmentations. To mitigate catastrophic forgetting, the student is also trained with supervised loss on \(\{x, Y\}\). The overall loss function is defined as:

$
    \mathcal{L} = \mathcal{L}_{hm}(Y, \hat{Y}) + w \cdot \mathcal{L}_{hm}\left(f(\mathcal{A}_s(x), \theta_s), \hat{Y}^u\right),
$

where \(\mathcal{A}_s\) denotes a strong augmentation function and \(\hat{Y}^u\) is the pseudo-label generated by the teacher model. Clearly, the quality of the pseudo-label \(\hat{Y}^u\) is crucial to the success of the student model. However, we empirically found that confidence-based pseudo-label filtering alone does not guarantee high-quality supervision. To address this, we leverage the inherent symmetry of CryoET data along the x, y, and z axes by applying flips in all three directions as weak augmentations. The teacher processes all four views (original and three flipped), and their outputs are averaged to produce a more reliable \(\hat{Y}^u\). Since the teacher is frozen during training, this multi-view inference introduces only minimal computational overhead.

\subsection{CryoET-specific DropBlock Augmentation}
\label{3_4}

In the co-training process, the choice of strong augmentations \(\mathcal{A}_s\) is pivotal. While strong augmentations such as RandAugment \cite{cubuk2020randaugment}, MixUp \cite{zhang2017mixup}, and CutOut \cite{devries2017improved} have proven effective in natural image domains, their direct application to CryoET is suboptimal due to the distinct characteristics of tomographic data.

\begin{figure}[h]
    \vspace{-5pt}
    \footnotesize
    \centering
\includegraphics[width=0.4\textwidth]{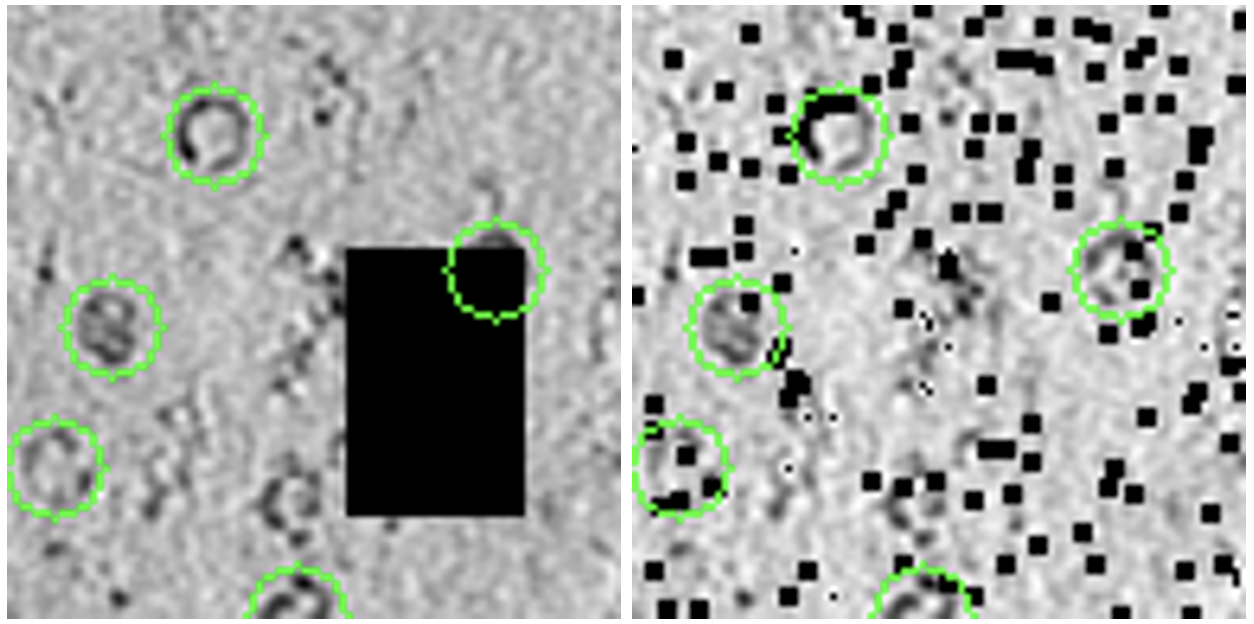}
    \caption{Left: CutOut augmentation. Right: CryoET-specific DropBlock augmentation.}
    \label{fig:interferance_case}
    \vspace{-10pt}
\end{figure}

In natural images, foreground objects are often large and sparse. CutOut, which removes random rectangular patches, mimics occlusions—a realistic challenge in object detection. In contrast, CryoET features small, densely packed particles. Applying CutOut often removes entire particles, depriving the model of critical training information. To address this, we adapt DropBlock~\cite{ghiasi2018dropblock} for CryoET data. Originally proposed as a regularization technique for CNNs, DropBlock randomly removes contiguous regions of feature maps during training. In our adaptation, we first sample a binary mask from a Bernoulli distribution with the same shape as the image input. We then apply max pooling with a small kernel(size=3) and a stride equal to half the kernel size to expand the dropped regions. Finally, we threshold the result to produce a binary mask that selectively occludes parts of particles while preserving sufficient structural context for effective learning. In our experiments, integrating this tailored DropBlock variant into \(\mathcal{A}_s\) significantly boosts performance in the semi-supervised setting.

\section{Experiment}

\subsection{Dataset}

We use the CZII dataset ~\cite{peck2024annotating} to evaluate our algorithm because it is the largest CryoET dataset and provides multi-class annotations. It includes center annotations for six particle categories, with 7 tomograms for training and 500 for testing. The significant imbalance stems from the high cost of annotating multiple particle types across many tomograms, which is infeasible for most biology labs~\cite{peck2024annotating}. To simulate this scenario, we randomly select 2 and 5 tomograms as the labeled training set \(\mathcal{D}^L\), 50 tomograms as the unlabeled set \(\mathcal{D}^U\), and another 50 tomograms as the test set \(\mathcal{D}^T\). Each tomogram has dimensions \(183 \times 650 \times 650\) with a voxel spacing of 10~\AA. The six particle classes have radii ranging from 60~\AA\ to 150~\AA, representative of most particle sizes observed in practice, with typical radii shown in Table \ref{tab:detailed_results}. The classes are apo-ferritin (AF), beta-amylase (BA), beta-galactosidase (BG), ribosome (Ribo), thyroglobulin (Thy), and virus-like particle (VLP).

\subsection{Implementation Details}

For training, we crop subtomograms of size \(90 \times 90 \times 90\) as input. Our asymmetric U-Net~\cite{ronneberger2015u} and ResNet-34~\cite{he2016deep} backbones are adapted with 3D operators. Specifically, we replace the initial \(7 \times 7 \times 7\) convolutional stem in ResNet with three successive \(3 \times 3 \times 3\) convolutions, following~\cite{he2019bag}. Models are trained using the Adam optimizer~\cite{kingma2014adam} on two NVIDIA A30 GPUs. Intensity augmentations are applied using the MONAI framework~\cite{cardoso2022monai}. During inference, we employ a sliding window strategy with 0.25 overlap, and apply test-time augmentation via three-axis flipping.

\begin{table}[t]
\footnotesize
\centering
\caption{Evaluation on CZII dataset. \textit{Supervised full} and \textit{Supervised} are considered as the upper and lower bound for the performance comparison. The best results are in \textbf{bold}.}
\vspace{-10pt}
\begin{tabular}{l|cc|cc}
\toprule
\multirow{2}{*}{\textbf{Method}} & \multicolumn{2}{c|}{\textbf{2 Labeled (4\%)}} & \multicolumn{2}{c}{\textbf{5 Labeled (10\%)}} \\
 & \textbf{F1@0.5} & \textbf{F1@0.75} & \textbf{F1@0.5} & \textbf{F1@0.75} \\
\midrule
\textit{Supervised full} & 0.611 & 0.684 & 0.611 & 0.684 \\
\textit{Supervised} & 0.508 & 0.562 & 0.569 & 0.615 \\
\midrule
DeepFindET ~\cite{moebel2021deep, harrington2024open} & 0.241 & 0.250 & 0.328 & 0.341 \\
VNet \cite{milletari2016v}  & 0.507 & 0.518 & 0.564 & 0.574 \\
Mean Teacher ~\cite{tarvainen2017mean} & 0.527 & 0.573 & 0.586 & 0.633 \\
\textbf{ours} & \textbf{0.561} & \textbf{0.612} & \textbf{0.607} & \textbf{0.653} \\
\bottomrule
\end{tabular}
\label{tab:seg_results}
\vspace{-10pt}
\end{table}

\subsection{Comparison with Other Methods}

We evaluate model performance using the F1 score. A predicted center is considered a true positive if it falls within 0.5 times the radius of the corresponding ground truth center; otherwise, it is treated as a false positive. We also report results under 0.75 radius threshold.

We compare our method against a range of baselines. First, we evaluate it against \textit{Supervised}, which is trained on \(\mathcal{D}^L\) with our asymmetric UNet and reweighted MSE loss. This comparison demonstrates how effectively our semi-supervised learning pipeline leverages unlabeled data. To verify the validity of the \textit{Supervised}, we also compare it with other supervised methods such as DeepFindET~\cite{moebel2021deep, harrington2024open} and VNet \cite{milletari2016v}. Notably, DeepFindET is specifically optimized for Cryo-ET particle picking. Additionally, we implement a classical semi-supervised learning method, Mean Teacher~\cite{tarvainen2017mean}, as a baseline. To illustrate the upper bound of performance, we include a \textit{Supervised full} baseline trained on both \(\mathcal{D}^L\) and \(\mathcal{D}^U\) with ground truth labels.

As shown in Table~\ref{tab:seg_results}, \textit{Supervised} significantly outperforms DeepFindET. We attribute this to the fact that the CZII dataset is the first moderately sized multi-class dataset in this domain. Previous methods like DeepFindET were primarily tuned on single-class or synthetic datasets, limiting their generalizability. Furthermore, our reweighted MSE loss proves effective for addressing sparse distributions. VNet is trained using the same protocol as the \textit{Supervised}; nonetheless, our asymmetric UNet still outperforms VNet, particularly on F1@0.75. This suggests that moderate downscaling in the UNet decoder enhances classification performance. 

The Mean Teacher model, initialized from the \textit{Supervised}, shows notable improvement over the \textit{Supervised}, confirming its ability to utilize unlabeled data. Our method, SemiETPicker, surpasses Mean Teacher by a large margin. When compared with the upper bound \textit{Supervised full}, SemiETPicker underperforms on F1@0.75 but performs comparably on F1@0.5. We attribute this to the fact that \textit{Supervised full} benefits from full access to ground truth labels. The similarity in F1@0.5 scores arises because both models face challenges in accurately localizing small particles, and F1@0.5 is more sensitive to localization accuracy.

\begin{table}[h]
\footnotesize
\centering
\caption{Evaluation results for all particles.}
\vspace{-4pt}
\begin{tabularx}{0.45\textwidth}{c|*{6}{X}}
\toprule
Proteins & AF & BA & BG & Ribo & Thy & VLP \\
\midrule
Radius & 60~\AA\ & 65~\AA\ & 90~\AA\ & 150~\AA\ & 130~\AA\ & 135~\AA\ \\
\midrule
\textit{Supervised} & 0.616 & 0.130 & 0.309 & 0.684 & 0.384 & 0.854 \\
SemiETPicker & 0.641 & 0.148 & 0.364 & 0.719 & 0.429 & 0.884 \\
\bottomrule
\end{tabularx}
\label{tab:detailed_results}
\end{table}

To examine the detailed performance of SemiETPicker across all particle types, we present the metrics of both SemiETPicker and the \textit{Supervised} in Table~\ref{tab:detailed_results}. One key observation is the varying levels of difficulty in detecting the six particle types. For example, the \textit{Supervised} achieves an F1@0.5 of 0.854 for VLP, while only reaching 0.130 for BA. Notably, SemiETPicker consistently improves upon the \textit{Supervised} baseline across all particles, regardless of whether the \textit{Supervised} performance is high or low.



\subsection{Ablation Studies and Efficiency Analysis}

\newcommand{\cmark}{\textcolor{green}{\Checkmark}}
\newcommand{\xmark}{\textcolor{red}{\XSolidBrush}}


\begin{table}[h]
  \centering
  \footnotesize
  \caption{Ablation study.}
  \label{tab:ablA}
  \vspace{-10pt}
  \begin{tabular}{ccc|c}
    \toprule
     MT & MV & DropBlock & F1@0.5\\ \midrule
     \xmark & \xmark & \xmark & 0.569\\
     \cmark & \xmark & \xmark & 0.586 \\
     \cmark & \cmark & \xmark & 0.595\\
     \cmark & \cmark & \cmark & \textbf{0.609}\\
    \bottomrule
  \end{tabular}
  \vspace{-10pt}
\end{table}

\begin{table}[h]
  \centering
  \footnotesize
  \caption{Efficiency analysis. Tested on one L40S GPU.}
  \label{tab:ablB}
  \vspace{-10pt}
  \begin{tabular}{c|cc}
    \toprule
     Methods & Time & F1@0.5\\ \midrule
     DeepFindET \cite{moebel2021deep} & 41s & 0.328 \\ \midrule
     VNet \cite{milletari2016v} & 2.66s & 0.564\\
     FPN \cite{lin2017feature} & 0.58s & 0.547\\
     Ours & 0.66s & \textbf{0.569}\\
    \bottomrule
  \end{tabular}
  \vspace{-10pt}
\end{table}

To validate the effectiveness of the modules we proposed, we design a series of ablation experiments. In Table~\ref{tab:ablA}, MT denotes Mean Teacher, MV denotes Multi-View pseudo labeling, and DropBlock refers to the CryoET-specific DropBlock data augmentation. As shown in the results, each of the three modules contributes to performance improvement, validating their effectiveness. Among them, the introduction of MT yields the largest gain of 0.017, demonstrating that this classic semi-supervised learning technique—composed of EMA updates and pseudo labeling—is also robust for the particle picking task. Our proposed Multi-View and DropBlock modules further improve the F1@0.5 by 0.023.

Another advantage of SemiETPicker is its efficiency. This efficiency stems from two design choices. First, our model outputs a heatmap, and the post-processing involves a simple max pooling operation, which can be efficiently accelerated by GPUs. As shown in Table ~\ref{tab:ablB}, DeepFindET is significantly slower than the other three methods. This is because DeepFindET adopts a segmentation-based design, requiring Connected Component Labeling (CCL) ~\cite{horn1986robot} as a post-processing step to extract particle centers. This operation runs on the CPU and is not well-suited for parallel computation, leading to a substantial bottleneck in the pipeline. Second, our asymmetric UNet design offers a favorable speed–accuracy trade-off. In Table~\ref{tab:ablB}, VNet and FPN \cite{lin2017feature} are trained and evaluated with the same pipeline as our supervised model, differing only in the backbone. This isolates the effect of postprocessing. FPN runs faster without skip connections but performs much worse. VNet, which predicts at input resolution, matches our accuracy but is far slower.

\section{Conclusion}

We present SemiETPicker, a fast and label-efficient semi-supervised framework for particle picking in CryoET tomograms. By integrating a lightweight heatmap-supervised detection model with a teacher-student co-training strategy, our method effectively exploits large-scale unlabeled data alongside sparse labeled data. The proposed multi-view pseudo labeling and a CryoET-specific DropBlock augmentation enable our framework to perform robustly on challenging low-SNR 3D data. Experimental results on the CZII dataset demonstrate significant improvements over supervised baselines, underscoring the promise of semi-supervised learning in advancing high-resolution structural analysis within cellular environments.

\vfill\pagebreak

\section{Compliance with ethical standards}
\label{sec:ethics}

This research study was conducted retrospectively using human subject data made available in open access. Ethical approval was not required as confirmed by the license attached with the open access data.

\section{Acknowledgments}
\label{sec:acknowledgments}

This work was supported by IMLS National Leadership Grant(LG-254883-OLS-23): Curating Very Large Biomedical Image Datasets For Librarian-In-The-Loop Deep Learning.



\begingroup
\footnotesize
\bibliographystyle{IEEEtran} 
\bibliography{strings,refs}
\endgroup

\end{document}